\documentclass[letterpaper, 10 pt, conference]{ieeeconf}
\IEEEoverridecommandlockouts

\usepackage{graphicx}
\usepackage{cuted}
\usepackage{cite}
\usepackage{subcaption}
\title{\LARGE \bf Hierarchical Unsupervised Topological SLAM}
\author{
    Ayush Sharma$^{*1}$, Yash Mehan$^{*1}$, Pradyumna Dasu$^{1}$, Sourav Garg$^{2}$, K. Madhava Krishna$^{1}$%
    \thanks{* Denotes Equal Contribution}%
    \thanks{$^{1}$ IIIT Hyderabad {\tt\small mkrishna@iiit.ac.in}}%
    \thanks{$^{2}$ University of Adelaide {\tt\small sourav.garg@adelaide.edu.au}}%
    \thanks{The authors thank MathWorks for their generous financial support.}
}
\begin{document}

\maketitle
\thispagestyle{empty}
\pagestyle{empty}
\begin{strip}
    \centering
    \includegraphics[scale=0.25]{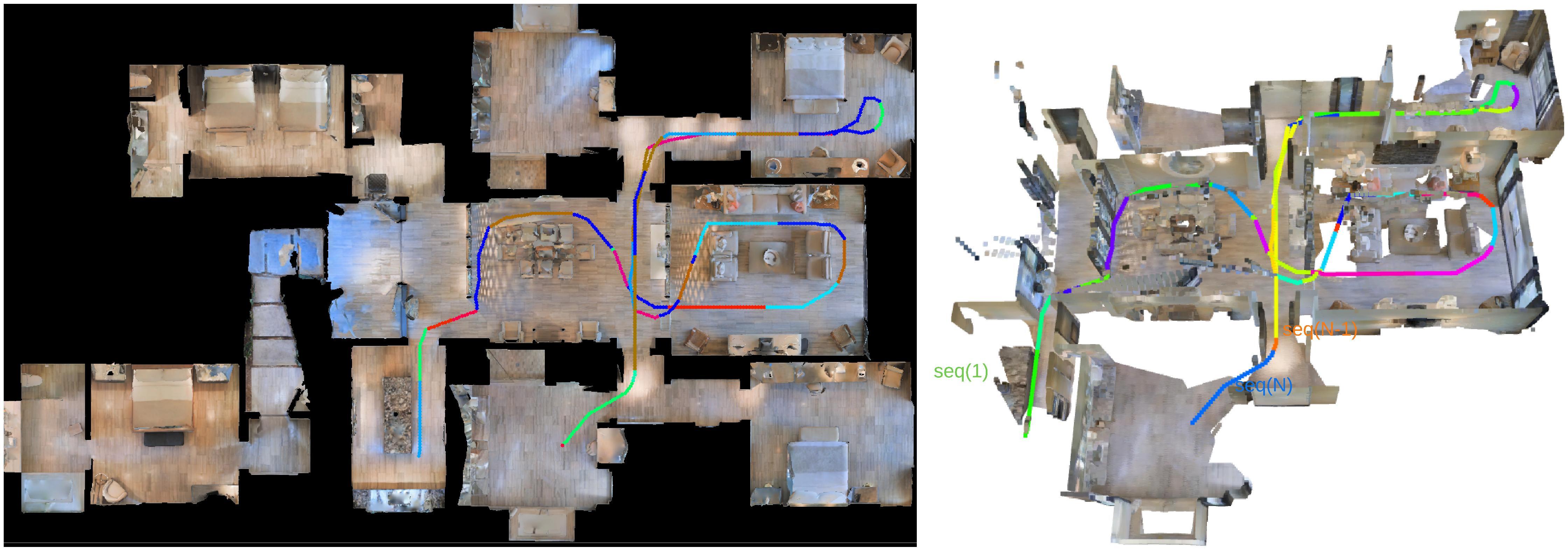}
    \captionof{figure}{ [Left] A top-down view of Matterport3D \cite{Matterport3D} scene, [Right] An embodied agent's traversal and sequences obtained from trajectory unsupervised segmentation.}
    \label{fig:mp3d_scene}
\end{strip}
\begin{abstract}

In this paper we present a novel framework for unsupervised topological clustering resulting in improved loop detection and closure for SLAM. A navigating mobile robot clusters its traversal into visually similar topologies where each cluster (topology) contains a set of similar looking images typically observed from spatially adjacent locations. Each such set of spatially adjacent and visually similar grouping of images constitutes a topology obtained without any supervision. We formulate a hierarchical loop discovery strategy that first detects loops at the level of topologies and subsequently at the level of images between the looped topologies. We show over a number of traversals across different Habitat environments that such a hierarchical pipeline significantly improves SOTA image based loop detection and closure methods.

Further, as a consequence of improved loop detection, we enhance the loop closure and backend SLAM performance. Such a rendering of a traversal into topological segments is beneficial for downstream tasks such as navigation that can now build a topological graph where spatially adjacent topological clusters are connected by an edge and navigate over such topological graphs.

\end{abstract}
\begin{figure*}
    \centering
    \includegraphics[width=1\linewidth]{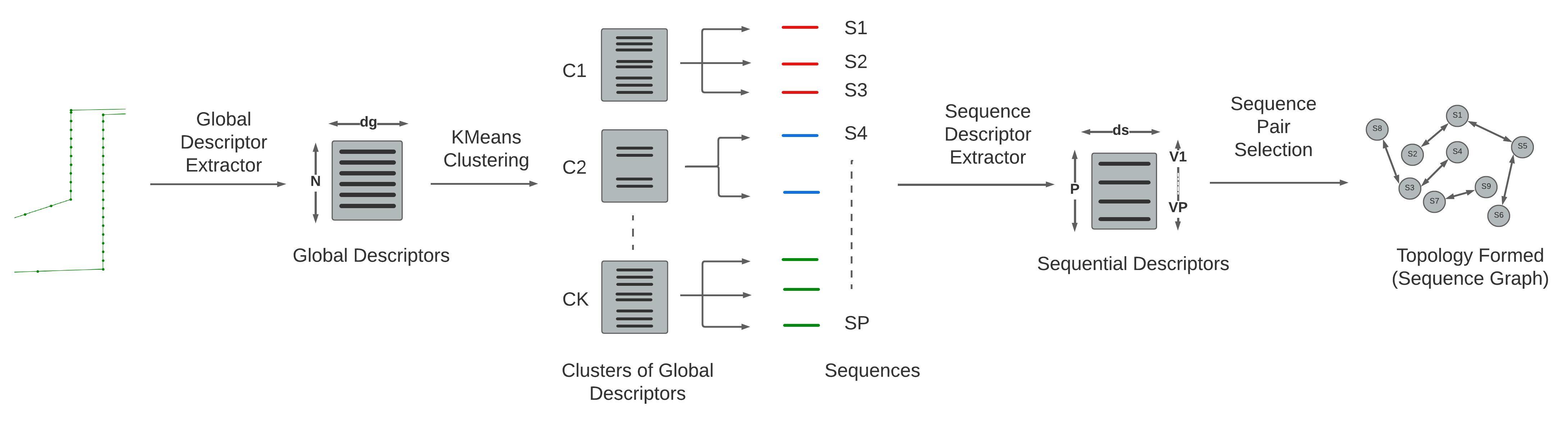}
    \captionof{figure}{Topology Formation Pipeline. Feature extraction, clustering then sequential descriptor extraction followed by pairing sequential descriptor based on their similarity score to output the topology formed(visualised as sequence graph). The sequences $S_{i}$'s will be having variable length i.e. number of frame will be different in different frames. The edges between two nodes in sequence graph, says that both node (sequences) belong to same topology. Here, $N$ and  $d_{g}$ denote total number of pose/frames and global descriptor dimension respectively. And, $P$, $V$ and  $d_{s}$ denote total number of sequences, sequence descriptor and it's dimension respectively.}
    \label{fig:pipeline}
\end{figure*}
\section{INTRODUCTION}
\label{sec:intro}

The SLAM problem has been widely studied in robotic and computer vision communities exploring various aspects of the problem. \cite{handbook, prob_robotics} discuss about early SLAM frameworks. The SLAM taxonomy includes classification based on sensing modality (monocular SLAM \cite{ORBSLAM, LSD_SLAM}, those that incorporate depth data \cite{KinectFusion}, and those that use LIDAR \cite{liosam2020shan}), incorporate multiple robots \cite{Tian2021KimeraMultiRD}, data-driven paradigms \cite{teed2021droid}, and those that incorporate semantics and objects \cite{Parkhiya2018ConstructingCM}. However, there have been very few approaches that integrate a topological understanding \cite{Puligilla2020TopologicalMF}.

Nonetheless, topological understanding is beneficial to both localization and mapping, as amply demonstrated in \cite{Fcil2019ConditionInvariantMP}. In this paper, we exemplify a novel unsupervised topological SLAM framework that segments the observations (images) accrued during a traversal into clusters. Each such cluster of images demarcates a topology (see Figures \ref{fig:mp3d_scene}, \ref{fig:traversal}) and can now be represented with a single representative embedding. The images that constitute a cluster are also obtained from spatially and temporally adjacent locations, and hence we call a topological embedding also as a sequence embedding.

As a direct consequence of segmenting a traversal into such sequence embeddings, we demonstrate improved loop detection and closures, especially so for sequences viewing the same scene but from a very different and disparate approach direction (see Figure \ref{fig:traversal}).

Specifically, the paper makes the following contributions:
\begin{enumerate}
    \item Proposes a novel, and one of the first such, framework for unsupervised hierarchical topological SLAM.
    \item We show enhanced loop detection and closure, exploiting the advantages offered by the hierarchical representation. We illustrate loop detection of sequences that observe the same topology or scene from disparate approach directions, making use of layered representations made possible due to the hierarchy.
    \item Quantitatively, we portray consistently higher Precision-Recall values vis-a-vis baselines \cite{DIR1, DIR2} that do not resort to sequential descriptor matching for loop detection.
    \item Further, we integrate our unsupervised topological SLAM framework with the popular RTABMAP to show improved loop detection and backend pose-graph optimization.
\end{enumerate}
\section{RELATED WORK}
\label{sec:related_work}
\subsection{Image Representation}
\label{subsec:img_repr}

Architectures like DBoW2 \cite{dbow2}, GeM \cite{DIR2}, NetVLAD \cite{netvlad}, and more recently, CosPlace \cite{berton2022rethinking} have been effectively demonstrated for global image representations for retrieval tasks. DBoW2 performs classical indexing and converts images into a bag-of-words representation, while building a hierarchical tree for approximating nearest neighbours in the image feature space and creating a visual vocabulary. Several recent works had kept focus on deep network methodologies \cite{DIR1, DIR2, netvlad}, especially improving pooling step for better utilisation of visual information present in an image. 

Recent work \cite{unsupervisedVPR} shows that \textit{K-STD} achieves better performance than the raw and standardized descriptors over the evaluated range of \textit{K}. Comparatively better result by \textit{K-STD} can be argued based on the exploitation of the clustering ability of global descriptors into meaningful clusters, i.e. visually similar images in one cluster and non-similar in separate clusters. Therefore, one default inherited property of global descriptors is to provide us better clustering (or grouping) of images, via any unsupervised clustering algorithms like KMeans and KMeans++.
\begin{figure*}
    \centering
    \includegraphics[width=1\linewidth]{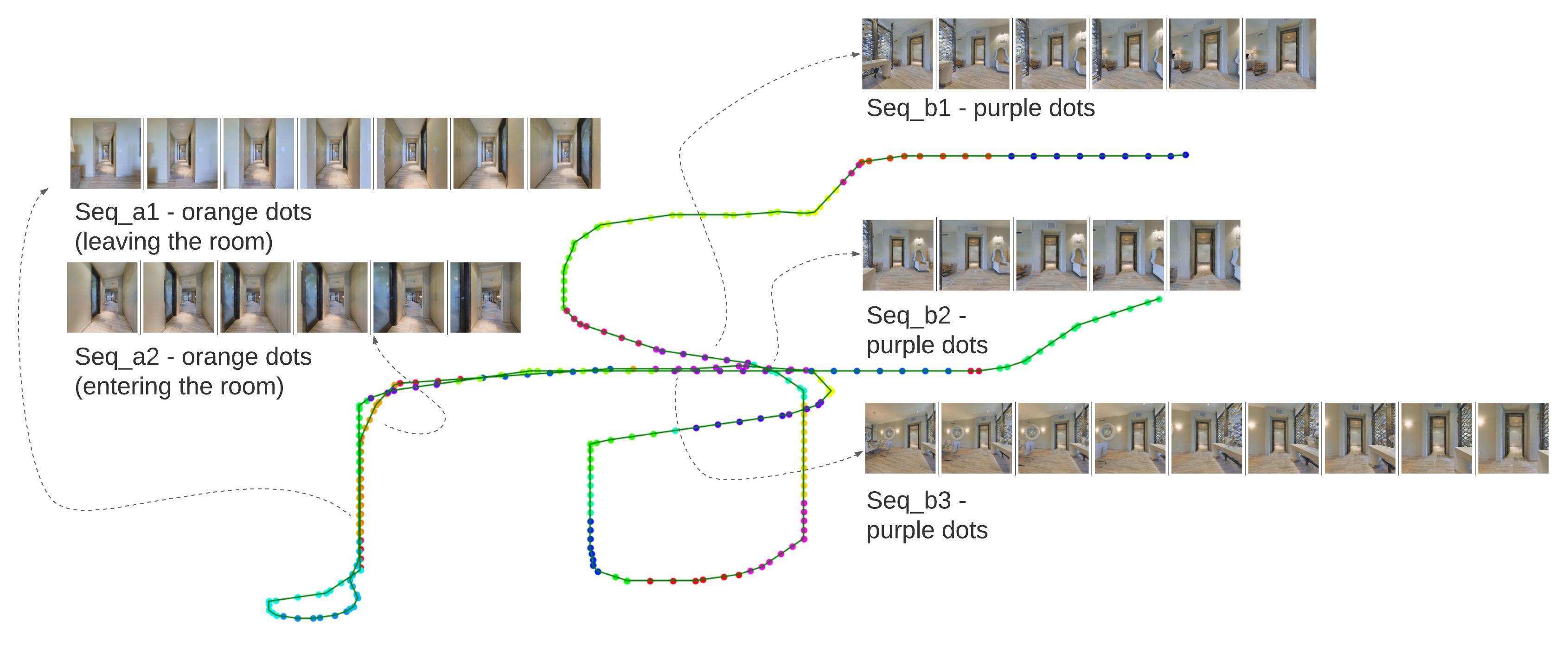}  
    \caption{Topology formation from agent's run. Figure visualises the segmentation of a traversal into sequences. Dots of a colour belong to one cluster. Sequences $Seq_{a1}$ and $Seq_{a2}$ belong to same cluster. So does $Seq_{b1}$, $Seq_{b2}$ and $Seq_{b3}$.}
    \label{fig:traversal}
\end{figure*}
\subsection{Sequence Representation}
\label{subsec:seq_repr}

Sequence-based place recognition has been extensively studied in the field of localization, with earliest approaches based on the post processing of a distance matrix computed by matching single image descriptor \cite{ho2007detecting, milford2012seqslam, arroyo2017you, vysotska2017relocalization, neubert2019neurologically}. A recent trend in this area has been to use sequential \textit{descriptors} to match places across reference and query traverses. 
\textit{Facil et al.} \cite{Fcil2019ConditionInvariantMP, vpr_survey_sunderhauf} used three basic techniques: concatenation of single image descriptors, fusion of the frame-level features with an FC layer, and integration over time of the single-image features via an LSTM network. Delta descriptors proposed in \cite{garg2020delta} provide an unsupervised method for sequential representation. SeqNet \cite{garg2021seqnet} proposed hierarchical sequential VPR which uses sequential descriptor based matching to guide single frame-based score aggregation. A more recent work \cite{mereu2022learning} proposes the categorization of architectures depending on the stage where the fusion mechanism is applied, i.e. late fusion like GeM+CAT, early fusion like Timesformer \cite{timesformer} and intermediate fusion like SeqNet \cite{garg2021seqnet} and SeqVLAD \cite{garg2021seqnet, mereu2022learning}. 
While sequence based approaches to VPR problem are generally superior to single image based approaches, use of sequential descriptors in particular reduces the cost of matching by incorporating temporal clues into the descriptor. This is because sequential descriptor methods summarize each sequence with a compact single vector and then perform the similarity search directly sequence-by-sequence.\\\\
However, a key issue persists in these sequential descriptor techniques: they are all demonstrated to be only effective for a \textit{fixed} sequence length setting. This is preset for each training run for both SeqNet \cite{garg2021seqnet} and SeqVLAD \cite{mereu2022learning}, thus requiring a new fixed sequence length model to be trained every time.  While the original SeqVLAD was not demonstrated for a variable length aggregation, our experiments show that it can be effectively used for sequence length unaware topology extraction and loop closures, refer \ref{subsec:exp_setup} for the same.

\section{METHODOLOGY}
\label{sec:method}

We first introduce the proposed pipeline of topology extraction. This is based first on clustering of image-level embeddings, followed by enforcing temporal connectivity on said clusters to obtain `sequences' from within those clusters. Then, we discuss a pose-based topology estimation and demonstrate our pipeline for the task of loop detection and closure for visual SLAM.
\subsection{Topology Formation}
\label{subsec:topology}

By topology we imply a collection of images whose individual embeddings are closer to the representative embedding of the entire collection vis-a-vis representative embeddings of other such collection. We also ensure that such collection of images are spatially close, as well as comprise of images viewing the same place from several different viewpoints, gathered potentially through multiple visits of that place during the agent's traversal. By the task definition, some of the observations would be temporally close, that is, consecutive image frames during the traversal. Our proposed topology formation (see Figure \ref{fig:pipeline}) is comprised of three steps: single image feature extraction, feature clustering to obtain sequences based on intra-cluster temporal connectivity, and finally, extraction of sequential descriptors which are the representative feature vectors or embeddings for that topology.
\begin{figure*}
    \begin{subfigure}{.5\textwidth}
        \centering
        \includegraphics[width=1\linewidth]{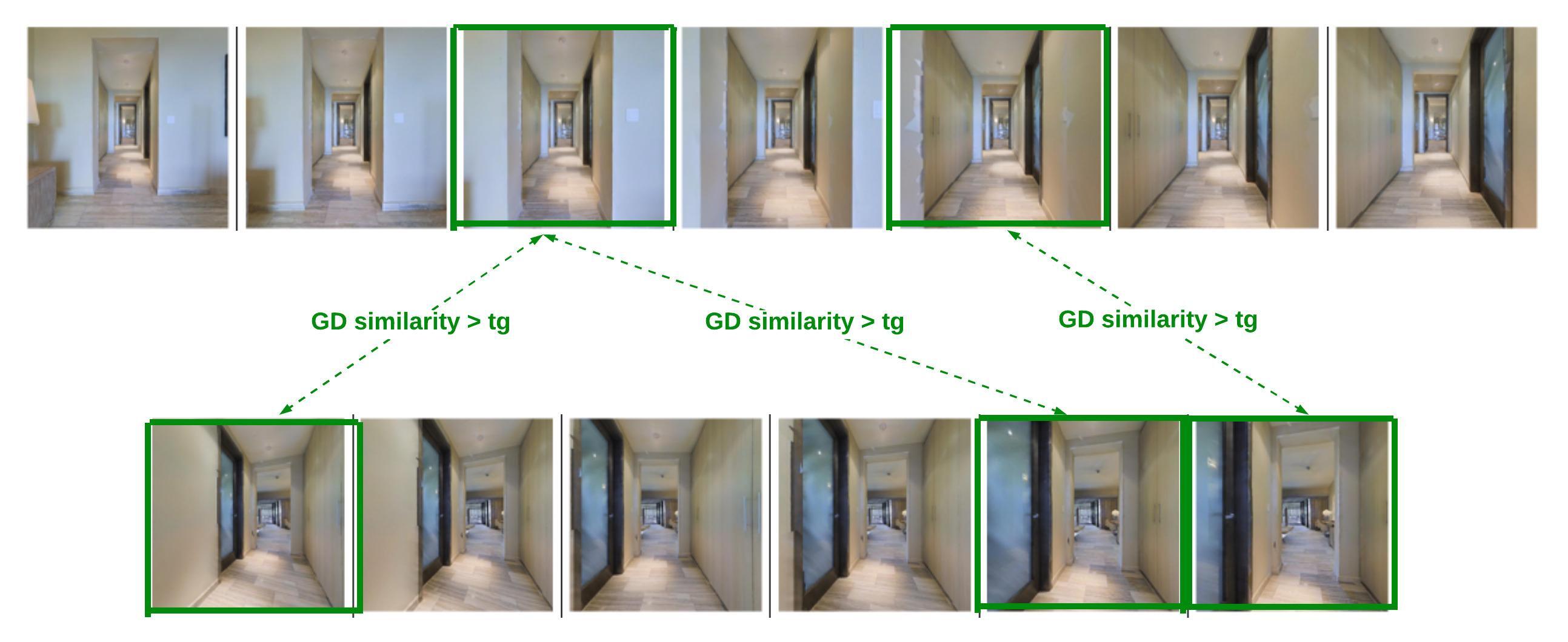}  
        \caption{Image matching between $Seq_{a1}$ and $Seq_{a2}$}
    \end{subfigure}
    \begin{subfigure}{.5\textwidth}
        \centering
        \includegraphics[width=1\linewidth]{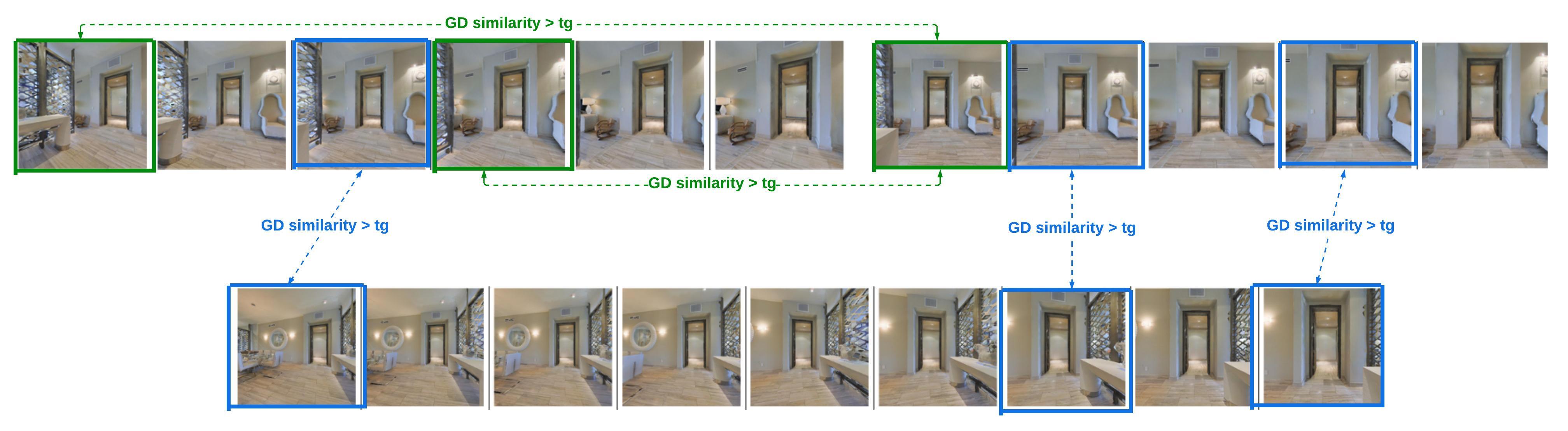}  
        \caption{Image matching between $Seq_{b1}$, $Seq_{b2}$ and $Seq_{b3}$}
    \end{subfigure}
    \caption{Intra-Topology Matching.  exhaustive image to image matching between selected sequence pair from the sequence graph to output the loop pairs is shown. Green pairs show loop pairs with visible overlap. Blue pairs show loop pairs having seemingly less visual overlap.}
    \label{fig:intra_topology_matching}
\end{figure*}
\subsubsection{Feature Extraction}
\label{subsubsec:feat_ext}

Given RGB images from the agent's traversal of the environment, say $I_{i}~(i~\in~[1,N])$, we extract global descriptors \textbf{$D_{i}$} for each image. Our method is agnostic to the choice of the global descriptor, thus both classical methods like DBoW2 \cite{dbow2} or more recent deep network based architectures like NetVLAD \cite{netvlad} and GeM \cite{DIR1, DIR2} can be used. We used the latter due to their demonstrated high performance; specifically, we used ResNet101+GeM \cite{DIR1, DIR2} for obtaining a 2048-dimensional compact vector embedding as a global descriptor for each image.
\subsubsection{Clustering}
\label{subsubsec:clustering}

In this step, the extracted global descriptors are clustered using KMeans algorithm. We used the elbow curve technique to decide a suitable \textit{K} (i.e., the number of centroids). This results in clusters with images that view the same place. These images would typically occur in the form of sequences, including those farther in timestamps due to multiple visits of the same place. At the same time, a cluster might comprise `outliers', that is, it could contain images that belong to a different place which is physically far apart from the majority cluster members. These \textit{K} clusters can be sub-divided into sequences, defined as a set of frames that are consecutive (that is, temporally connected). 

After clustering the global descriptors $D_{i}$, we obtained clusters $C_{k}~(k~\in~[1,K])$. We sub-divide each of these clusters to obtain sequences $S_{jk}~(j~\in~[1,P_k])$, where $P_k$ refers to number of sequences in $k^{th}$  cluster and each such sequence will belong to exactly one of the $K$ clusters (see Figures \ref{fig:pipeline}, \ref{fig:traversal}). Also, note that the number of frames will vary across sequences $S_{jk}~(j~\in~[1,P_k])$, i.e. sequences of different lengths will be there.
\subsubsection{Sequential Descriptor}
\label{subsubsec:seq_desc}

From the sequences obtained from the previous step, $S_{jk}$ for each cluster $k$, we compute their corresponding sequential descriptors $V_{jk}$. This is achieved through SeqVLAD \cite{mereu2022learning}, which we slightly adapt to obtain embeddings for \textit{variable} length sequences, even though it is trained on fixed length sequences as described in section \ref{subsec:seq_repr}. 

While we obtained sequences as members of the clusters formed through single images, a more informed representation of a place is achieved through the sequential descriptors. Thus, we can now compare sequential descriptors \textit{both} within and across clusters to finally decide the topologies. For this purpose, we use cosine similarity $s_{pp'} \in [-1,1]$ to compare two sequence descriptors $V_{p}$ and $V_{p'}$ where $p,p' \in [1,\sum_kP_k]$ refer to the indices of sequences regardless of their cluster $k$. If $s_{pp'} \ge t_s$, then sequences $S_{p}$ and $S_{p'}$ belongs to the same topology. 

Visualisation of the topology so formed can be done in graphical way, where nodes represent sequences and an edge between two nodes implies that they belong to the same topology. We can refer to the obtained graph as a sequence graph or sequence based topological graph (see Figure \ref{fig:pipeline}).
\subsection{Intra-Topology Matching}
\label{subsec:intra-topology}

 Following Section \ref{subsec:topology}, one can get a sequence based topological graph (sequence graph) for an agent's traversal in an environment. Then, intra topological matching can be done as follows: for all the sequence pairs having an edge between them in the sequence graph, an exhaustive image to image (first sequence $S_{jk}$ to second sequence $S_{j'k}$) global descriptor ($D_i$) matching is performed. The image pairs having a global descriptor similarity score above the threshold $t_g$ are detected as loop pairs (see Figure \ref{fig:intra_topology_matching}). 
\subsection{SLAM Pipeline}
\label{subsec:slam_pipeline}

Formally, given a visual traversal, the loop detection task is to compute correct loop pairs within a radius, say 5 m. Vision based loop detection task utilises VPR (visual place recognition) \cite{vpr_survey_sunderhauf} solutions as core modules. The topology formation technique along with the intra-topological matching forms a two-stage hierarchical loop detection pipeline, which detects loop closure candidates at the sequence and the image levels respectively. We propose a SLAM pipeline whose frontend comprises of this hierarchical loop detection pipeline and a robust local feature matching pipeline like OriNet \cite{Laguna2019KeyNetKD, AffNet2017, HardNet}, with the backend optimisation handled by G2O \cite{g2o}. One widely accepted SLAM framework, RTABMAP \cite{labbe2019rtab}, performs the same task but employs DBoW2 based loop detection system with additional loop hypothesis filtering modules before local feature matching step.

Later, we show that augmenting RTABMAP with our proposed hierarchical system's loop pairs provides a better final trajectory optimisation with G2O \cite{g2o} (Section \ref{subsec:rtabmap}).
\section{EXPERIMENTATION AND RESULTS}
\label{sec:exp_results}
\subsection{Setup}
\label{subsec:exp_setup}

The dataset used was Matterport3D~\cite{Matterport3D} and Habitat-sim ~\cite{habitat19iccv} for simulating an embodied agent traversing the environment. 12 environments with 2 trajectories each were chosen at random for training of the sequence descriptor extractor. The method of  \textbf{III A2} was employed to extract sequences for each trajectory. The obtained training sequences, culled to a static length(5 for experiments in this paper) were employed during training SeqVLAD and one of late fusion model~\cite{mereu2022learning} ResNet18l3+GeM+CAT. Small sequence length is in response to fast visual change in indoor setting as compared to outdoor scenarios. Finally, our models have been trained following same training method, as defined in \cite{mereu2022learning}.

During inference, given the agent's traversal, method as in  \textbf{III A2} was employed to extract variable length sequence. SeqVLAD is an intermediate fusion model~\cite{mereu2022learning} and extended version of NetVLAD \cite{netvlad}. VLAD \cite{netvlad, mereu2022learning} layer output is independent of height, width and depth dimensions of the feature maps obtained after convolution layers, but not of the sequence dimension. Keeping this in mind, we infer one sequence at a time through SeqVLAD i.e. inference batch size one that enabled us to extract fixed length representations for variable length sequences.

\subsection{Augmenting RTABMAP}
\label{subsec:rtabmap}

Unsupervised SLAM indoors is a challenging yet promising avenue due to presence of repeating scenes and potential of opposite view loop pairs while re-traversing paths. The ability of our system to capture loop pairs having extreme viewpoints difference has been highlighted in figure \ref{fig:histogram}, which RTABMAP \cite{labbe2019rtab}  and DBoW2 fails to detect.

\begin{figure}[ht]
    \begin{subfigure}{0.49\linewidth}
        \centering
        \includegraphics[width=\linewidth]{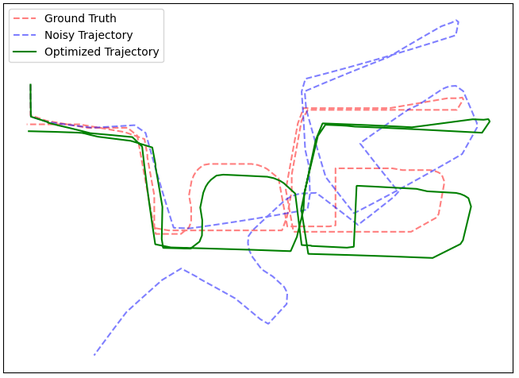}  
        \caption{Run 1 Augmented loop pairs}
    \end{subfigure}
    \begin{subfigure}{0.49\linewidth}
        \centering
        \includegraphics[width=\linewidth]{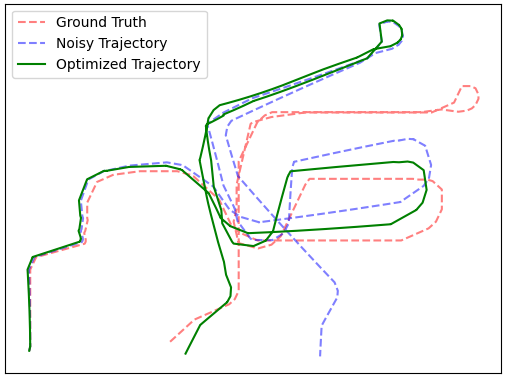}
        \caption{Run 2 Augmented loop pairs}
    \end{subfigure}
    \begin{subfigure}{0.49\linewidth}
        \centering
        \includegraphics[width=\linewidth]{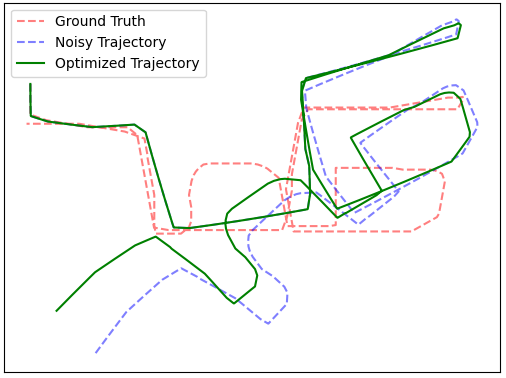}  
        \caption{Run 1 RTABMAP loop pairs}
    \end{subfigure}
    \begin{subfigure}{0.49\linewidth}
        \centering
        \includegraphics[width=\linewidth]{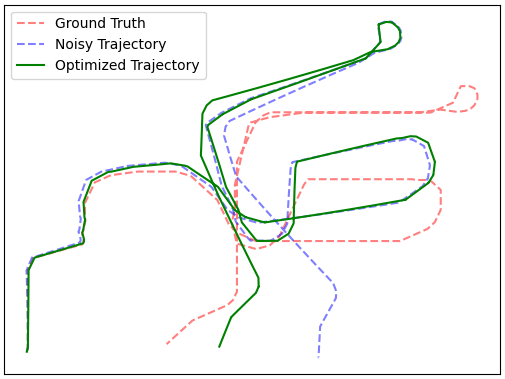}  
        \caption{Run 2 RTABMAP loop pairs}
    \end{subfigure}
    \begin{subfigure}{0.49\linewidth}
        \centering
        \includegraphics[width=\linewidth]{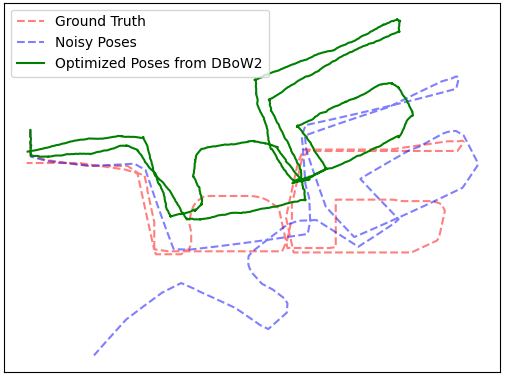}  
        \caption{Run 1 DBoW2 loop pairs}
    \end{subfigure}
    \begin{subfigure}{0.49\linewidth}
        \centering
        \includegraphics[width=\linewidth]{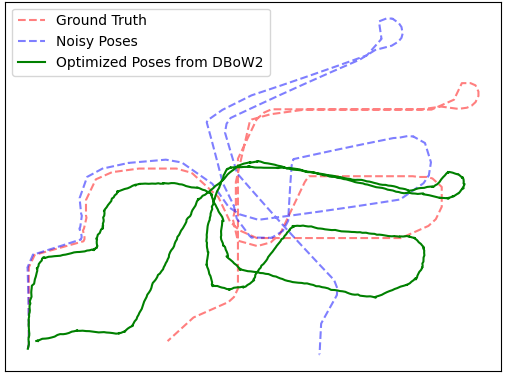}  
        \caption{Run 2 DBoW2 loop pairs}
    \end{subfigure}
    \caption{Pose-graph optimisation results using G2O. We obtained the noisy trajectory by adding Gaussian noise to the ground truth odometry information. External noise has been added to account for drift accumulated in the real world dataset. The subfigures (a) and (b) compare optimization of noisy trajectories based on augmented loop pairs, (c) and (d) show optimization of noisy trajectories based on RTABMAP's native loop pairs only. (e) and (f) show optimisation based on DBoW2's loop pairs. RTABMAP + Our augmented loop pairs enhance the optimization as compared to RTABMAP's native or DBoW2's loop pairs. This observation is consistent across numerous possible noise having different mean and variance.}
    \label{fig:optimization}
\end{figure}

\begin{figure}[ht]
  \centering
  \includegraphics[width=1\linewidth]{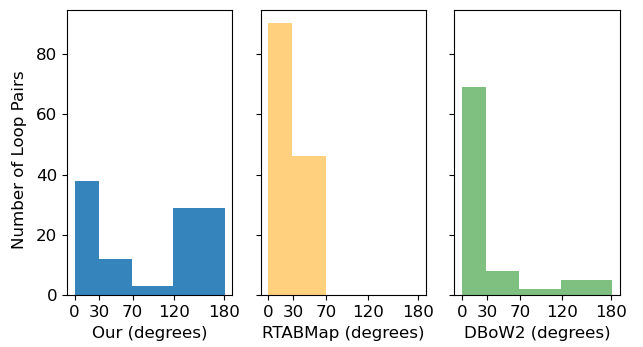}  
  \caption{Ground truth rotational difference between candidates of a loop pair in degrees for different loop detection pipeline. Detection of obtuse and opposite view loop pairs is significantly improved due to our formulation as compared to RTABMAP's and DBoW2's loop detection. This compensates any reduction observed in the proposed system while dealing with the loop pairs  that have very close or similar viewing angle. The improved backend optimization due to the augmented framework confirms this in Figure \ref{fig:optimization} and Table \ref{tab:ape}.}
  \label{fig:histogram}
\end{figure}

Post detection of opposite view pairs by our proposed method, OriNet \cite{Laguna2019KeyNetKD, AffNet2017, HardNet}, accepted for robustness to aggressive viewpoint changes, is employed for feature matching. Subsequently the frame transform between the loop pairs is passed to the backend optimizers. Figure \ref{fig:optimization} exhibits enormous boost in backend optimization of the trajectory post augmentation of loop pairs. The same can be seen in Table \ref{tab:ape} wherein the mean Absolute Pose Error of augmented loop pairs is significantly better than RTABMAP's native or DBoW2's detected loops.
\begin{table}[hbtp]
    \caption{Absolute Pose Error (APE) of trajectories w.r.t ground truth trajectories.}
    \begin{center}
        \resizebox{\columnwidth}{!}{
        \begin{tabular}{|c|c|c|c|c|c|c|}
            \hline
            \multicolumn{1}{|c|}{\textbf{ APE w.r.t Ground Truth }} &\multicolumn{3}{|c|}{\textbf{run1}} &\multicolumn{3}{|c|}{\textbf{run2}} \\
            \hline 
            \textbf{\textbf{Trajectory}} &\textbf{\textit{min}} &\textbf{\textit{mean}} &\textbf{\textit{max}} &\textbf{\textit{min}}&\textbf{\textit{mean}}  &\textbf{\textit{max}}   \\
            \hline
            Noisy & 0.011 & 1.715 & 6.457 & 0.224 & 0.930 & 3.576 \\ \hline
            Optimisd with DBoW2 loop pairs  & 0.084 & 0.956 & 2.475 & 0.127 & 0.782 & 1.77 \\ \hline
            Optimised with RTABMAP native loop pairs & 0.020 & 1.289 & 5.190 & 0.189 & \textbf{0.649} & 1.689 \\ \hline
            Optimised with RTABMAP+Our loop pairs & \textbf{0.018} & \textbf{0.312} & \textbf{0.757} & \textbf{0.08} & 0.69 & \textbf{1.673} \\ \hline
        \end{tabular}
        }
        \label{tab:ape}
    \end{center}
\end{table}
\subsection{Performance Analysis}\label{AA}
\label{subsec:perf_analysis}

A typical global descriptor based loop detection system considers every frame as the query and all the previous frames (except a temporally adjacent ones) as the reference set. To filter the detected loop pairs, all retrieved images which have their descriptor similarity score at-least some threshold  $t_g$ is considered. We make use of ResNet101+GeM descriptors for baseline comparison with our proposed hierarchical system.  In addition, we make baseline comparison with DBoW2 based loop detection system as a representative for BoW (bag of words) class of typical classical descriptor system.
\subsubsection{Precision and Recall}
\label{subsec:precision_recall}

\begin{figure}[ht]
    \centering
    \includegraphics[width=0.7\linewidth]{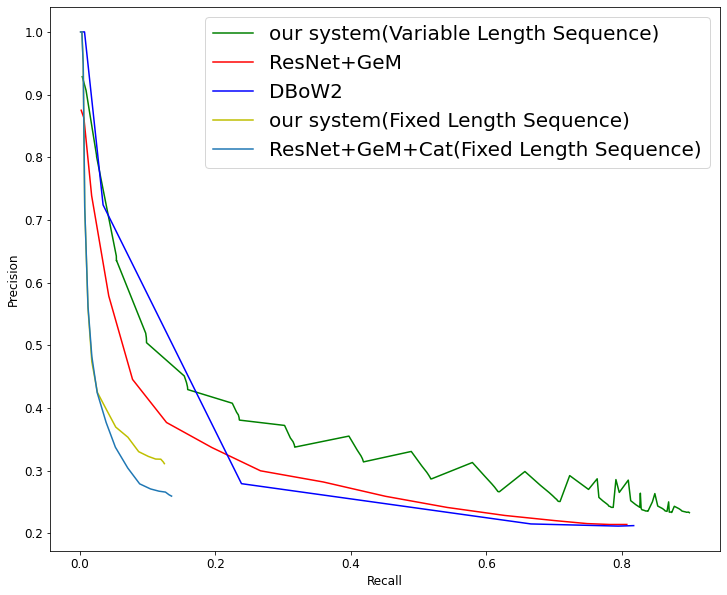}  
    \caption{PR-curves for loop pairs from different systems}
    \label{fig:pr_curve_variable_length}
\end{figure}

For loop detection system, we define recall as percentage of correct predicted loop pairs by the system out of all possible correct loop pairs, and precision as percentage of correct predicted loop pairs out of total predicted loop pairs. PR curves (see Figure \ref{fig:pr_curve_variable_length}) show that the proposed method is typically better for every pair of precision-recall values as the respective thresholds for the methods are varied vis a vis the ResNet+GeM method. The proposed method is on-par with the BoW methods but specifically is able to recall better loop pairs with significant disparity in viewing angles (see Figure \ref{fig:histogram}) 
\subsubsection{PR curve for fixed sequence length based loop detection systems}
\label{subsec:pr_curve}
In Figure \ref{fig:pr_curve_variable_length}, we showcase the advantages of SeqVLAD inherent ability to handle variable length sequences at inference time. When SeqVLAD is forced to use fixed length clusters at inference time the recall drops considerably whereas variable length SeqVLAD has a much better recall. Therefore, proposed hierarchical system faces significant decrement in the performance when constrained to fixed length sequences for the results in this paper. 
\section{CONCLUSION AND FUTURE WORK}
\label{sec:conclusion}

The aim of this work is to show that a clever use of image level and sequence level information for multi-stage hierarchical loop detection augments SLAM frameworks' capability for detecting loop closures and showing significant improvement over deep global descriptor methods.

Our work detects loop closures between perspective images of the same place with a wide shift in the viewpoint, where the classical methods fail to keep up. This opens up new avenues for improving the pose-graph optimization in SLAM frameworks that utilize methods which rely on the agent revisiting a scene in similar viewpoints.

\bibliographystyle{IEEEtran}
\bibliography{ref}

\end{document}